\documentclass[10pt]{extarticle} 

\usepackage[utf8]{inputenc}
\usepackage[T1]{fontenc}
\usepackage{hyperref}
\usepackage{url}
\usepackage{amsmath}
\usepackage{fancyhdr}
\usepackage[utf8]{inputenc}
\usepackage{booktabs}
\usepackage{amsfonts}
\usepackage{nicefrac}
\usepackage{microtype}
\usepackage{graphicx}
\usepackage{natbib}
\usepackage{doi}
\usepackage{tikz}
\usepackage{adjustbox}
\usepackage{enumitem}
\usepackage{listings}
\usepackage{color} 
\usetikzlibrary{shapes.geometric, arrows.meta, positioning}

\tikzstyle{startstop} = [rectangle, rounded corners, minimum width=3.5cm, minimum height=1cm,text centered, draw=black, fill=gray!10]
\tikzstyle{process} = [rectangle, minimum width=3.5cm, minimum height=1cm, text centered, draw=black, fill=blue!5]
\tikzstyle{arrow} = [thick, ->, >=Stealth]

\usepackage[margin=1in]{geometry}

\title{Text-to-Speech System for Meitei Mayek Script}

\author{
  Irengbam Gangular Singh \\
  \And
  Wahengbam Nirvash Singh \\
  \And
  Khumanthem Lanthoiba Meitei \\
  \And
  Paikhomba Oinam \\
}

\hypersetup{
pdftitle={Text-to-Speech System for Meitei Mayek Script},
pdfsubject={cs.CL, cs.LG},
pdfauthor={Irengbam Gangular Singh, Wahengbam Nirvash Singh, Khumanthem Lanthoiba Meitei, Paikhomba Oinam},
pdfkeywords={Text-to-Speech, Manipuri, Meitei Mayek, Tacotron2, HiFi-GAN, Low-resource language, Phoneme mapping},
}

\begin{document}

\begin{center}
    
    \vspace{1em}
    {\Large\bfseries Text-to-Speech System for Meitei Mayek Script\par}
    \vspace{1em}
    \vspace{2em}
    
    \begin{tabular}{ccc}
        \textbf{Irengbam Gangular Singh} & \textbf{Wahengbam Nirvash Singh} & \textbf{Khumanthem Lanthoiba Meitei} \\
        B.Tech, CSE & B.Tech, CSE & B.Tech, CSE \\
        NIT Manipur & NIT Manipur & NIT Manipur
    \end{tabular}
    
    \vspace{2em}
    
    \textbf{Paikhomba Oinam} \\
    B.Tech, CSE \\
    NIT Manipur
\end{center}

\begin{abstract}
This paper presents the development of a Text-to-Speech (TTS) system for the Manipuri language using the Meitei Mayek script. Leveraging Tacotron~2 and HiFi-GAN, we introduce a neural TTS architecture adapted to support tonal phonology and under-resourced linguistic environments. We develop a phoneme mapping for Meitei Mayek to ARPAbet, curate a single-speaker dataset, and demonstrate intelligible and natural speech synthesis, validated through subjective and objective metrics. This system lays the groundwork for linguistic preservation and technological inclusion of Manipuri.
\end{abstract}

\section{Introduction}
Language is not only a tool for communication but also a constituent of cultural
 heritage and identity. With the rapid development of digital communication and
 artificial intelligence, speech technologies like Text-to-Speech (TTS) have come to
 play central roles in advancing accessibility, improving human-computer interaction,
 and preserving linguistic diversity. However, TTS system development has, to a
 great extent, focused on a narrow selection of popular languages and Latin scripts,
 without giving attention to numerous indigenous languages and writing systems.
 Manipuri, officially one of India’s 22 scheduled languages, is spoken predominantly
 in the state of Manipur and also in communities in surrounding regions. Historic
 writing of Manipuri has taken place in Meitei Mayek, a historic script native to the
 region which saw a resurgence following decades of replacement by Bengali script
 use. It has extremely low visibility in modern-day computational linguistics and
 speech technology in spite of cultural and linguistic pertinence.
 This paper presents the design and development of a TTS system for Manipuri in the
 Meitei Mayek script. By combining deep learning-based TTS architectures such as
 Tacotron 2 and HiFi-GAN with a properly designed phoneme mapping to ARPAbet.
 This work attempts to fill a critical technological gap and make a
 contribution towards the digital empowerment of the Manipuri-speaking community.

\newpage
\section{Literature Review}

Text-to-speech (TTS) systems convert written text into spoken language, enabling applications such as voice assistants, audiobooks, and language learning. In recent years, deep‐learning TTS models have achieved near-human naturalness for well-resourced languages\cite{shen2018natural}\cite{kong2020hifi}.However, underrepresented and low-resource languages often lack the large, high-quality speech corpora that these models require \cite{anand2024elaichi}
\cite{devi2021manito}.This disparity means that speakers of many languages – including Manipuri – have limited access to state-of-the-art speech technology. Developing TTS for such languages is important both for technological inclusion and for preserving linguistic diversity; indeed, robust speech technologies can help prevent the loss of endangered languages \cite{devi2021manito}
\cite{anand2024elaichi}

Traditional TTS pipelines. Before the neural era, TTS systems typically comprised multiple hand-engineered modules (text analysis, phoneme duration modeling, prosody prediction, and a vocoder)\cite{wang2017tacotron} \cite{debnath2022low}.Constructing these components required extensive domain expertise.\cite{wang2017tacotron}.Parametric and concatenative TTS engines also tended to sound muffled or robotic 

Tacotron (Wang et al., 2017). Wang et al. introduced Tacotron, one of the first fully end-to-end neural TTS models \cite{wang2017tacotron}. Tacotron replaces the modular pipeline with a single sequence-to-sequence network: given pairs of raw text and audio, it learns to map character embeddings directly to intermediate acoustic features (spectrograms), and then uses a simple vocoder to synthesize audio.\cite{wang2017tacotron}.It was substantially faster at inference than autoregressive sample-level models \cite{wang2017tacotron}.

Tacotron2 (Shen et al., 2018). Tacotron2 refines the end-to-end approach by splitting it into two neural stages: a text-to-spectrogram model and a neural vocoder \cite{shen2018natural}. Tacotron2 achieved a ground-breaking MOS of 4.53, nearly indistinguishable from recorded speech (MOS 4.58) \cite{shen2018natural}. 

Low-resource TTS and transfer learning (Debnath et al., 2022). Debnath et al. (2022) present one of the few examples in the Indian context: they built an end-to-end Sanskrit TTS by fine-tuning an English-pretrained Tacotron2 on a small Sanskrit corpus \cite{debnath2022low}.Using only about 2.5 hours of Sanskrit speech, Debnath et al. were able to synthesize intelligible, natural-sounding Sanskrit. The final system (Tacotron2 + WaveGlow vocoder) achieved a subjective MOS of 3.38, which is impressive given the extremely limited data \cite{debnath2022low}.This work highlights that 
(a) transfer learning from a resource-rich language can bootstrap TTS for a related low-resource language, and 
(b) using a parallel neural vocoder like WaveGlow can provide decent quality even with little fine-tuning. The Sanskrit example is directly relevant to Manipuri: it suggests that similar strategies (pretraining on another language, then fine-tuning Tacotron2) could help overcome the data shortage.

Neural vocoders – HiFi-GAN vs. WaveGlow. Modern TTS systems often pair Tacotron-like acoustic models with neural vocoders to convert spectrograms to audio. Early work used Griffin-Lim or WaveNet, but these were slow or complex. WaveGlow (Prenger et al., 2019) improved efficiency by using flow-based generative modeling \cite{kong2020hifi}, and it was used in Debnath et al.’s Sanskrit TTS. More recently, Generative Adversarial Network (GAN)-based vocoders like HiFi-GAN (Kumaran et al., 2020) have demonstrated even better performance. The HiFi-GAN authors report that their model achieves higher perceptual quality (MOS) than both WaveNet and WaveGlow \cite{kong2020hifi}.For example, a tiny HiFi-GAN (only 0.92M parameters) still outperformed larger models, and HiFi-GAN synthesized “human-quality” speech at speeds of over 3.7 MHz on a single GPU \cite{kong2020hifi}. In practical terms, this means HiFi-GAN is faster and more parameter-efficient than WaveGlow, while matching or exceeding its audio fidelity. Many recent TTS systems therefore adopt HiFi-GAN as the vocoder of choice. In our work, we also use HiFi-GAN for waveform generation, expecting its efficiency and quality to benefit Manipuri TTS.

\subsection{Research Gaps}
Despite these advances, several gaps remain for Manipuri TTS:

1. Scarcity of Meitei Mayek data. There is no publicly available high-quality TTS corpus in Manipuri’s native script. Existing resources are limited and often not at studio quality. For example, the largest reported Manipuri speech collection is a telephonic ASR dataset (approx. 100h from 300+ speakers) \cite{patel2018development};  even this has narrowband audio and mixed speakers. No open dataset of recorded single-speaker Manipuri (Meitei Mayek) has been released. This contrasts sharply with languages like English or Hindi, for which many hours of single-speaker data exist. The lack of data hampers training Tacotron2 and other neural TTS models from scratch.

2. Tonal phonology is neglected. Manipuri is a tonal language with at least two distinctive tones (level and falling)\cite{devi2021manito}. These tones can change word meaning, and in Meitei Mayek the falling tone is explicitly marked with a diacritic \cite{devi2021manito}. However, most TTS research (including works cited above) has focused on non-tonal languages or has not explicitly modeled tone. For example, Tacotron2 and its variants have been developed for English, Spanish, and some Indian languages (e.g. Hindi, Gujarati) that are essentially non-tonal. There is little prior work on how to incorporate tone features into a TTS model. We must therefore extend existing frameworks to capture Manipuri’s tonal contrasts – for instance by conditioning the model on tone markers or pitch features.

3. Lack of prior Manipuri TTS studies. To our knowledge, very few TTS systems have been built for Manipuri or closely related Tibeto-Burman languages. Most Indian language TTS research has targeted Indo-Aryan (Hindi, Marathi) or Dravidian (Tamil, Telugu) languages with Brahmi-derived scripts. Even Sanskrit (an Indo-Aryan language) received almost no end-to-end TTS work until recently \cite{kumar2023towards}.While a project like AI4Bharat’s Indic-TTS mentions Manipuri (among 13 languages)\cite{kumar2023towards},  it is unclear whether that work uses the native script or addresses tone. In any case, the lack of published results on Manipuri TTS underscores the novelty of this effort. Both the dearth of data and the scarcity of modeling techniques tailored to Manipuri motivate our targeted research.

\section{Study Methodology}

This chapter outlines the stepwise methodology adopted for developing a Text-to-Speech (TTS) system for the Meitei Mayek script. The process is structured into multiple phases, starting from data acquisition to speech synthesis, and is visualized in 
Figure . Each phase is described in detail as follows.

\vspace{0.5em}
\textbf{Text Data Collection: }
The development process begins with manually writing sentences in the Meitei Mayek script. These texts are carefully composed by native speakers to ensure grammatical accuracy, script correctness, and coverage of diverse linguistic contexts. This manually curated corpus serves as the textual input for subsequent TTS stages.

\vspace{0.5em}
\textbf{Audio Recording: }
Simultaneously, corresponding audio samples are recorded by a native speaker proficient in Meitei Mayek. Care is taken to ensure high-fidelity recordings, consistent pronunciation, and minimal background noise, thus ensuring the audio data's quality and reliability.

\vspace{0.5em}
\textbf{Audio Preprocessing: }
The collected audio samples undergo preprocessing to optimize their suitability for model training. This includes: 

a. Silence Removal: Unwanted pauses at the beginning and end of each audio clip are eliminated.

b. Volume Normalization: Audio clips are standardized to maintain consistent loudness across the dataset.

These steps ensure the homogeneity of the audio data, which is critical for effective model learning.

\vspace{0.5em}
\textbf{Text Normalization: }
Text normalization aims to convert text into a canonical form. Tasks such as expanding abbreviations, standardizing numerical expressions, and unifying spelling variants are undertaken. Normalization ensures that the text is coherent and suitable for phoneme-level processing.

\vspace{0.5em}
\textbf{Phoneme Conversion: }
After normalization, the text is transcribed into phonemes.An adapted ARPAbet phoneme set is employed to encode the pronunciation. This phonetic representation bridges the gap between orthography and pronunciation, allowing the model to better learn the mapping from text to speech.

\vspace{0.5em}
\textbf{Text-Audio Alignment and Pairing: }
Phoneme sequences are then aligned precisely with their corresponding audio recordings. This step involves manual verification to ensure that the timing and content match accurately, forming a robust paired dataset for model training.

\vspace{0.5em}
\textbf{Model Training: Tacotron 2 Fine-Tuning: }
The paired dataset is used to fine-tune a Tacotron 2 model, originally designed for end-to-end speech synthesis. Tacotron 2 learns to predict mel-spectrograms from sequences of phonemes. During this phase:

\begin{enumerate}[label=\alph*.]
    \item \textbf{Text/Audio Error Correction:} Errors in the dataset are identified and rectified through manual intervention.
    \item \textbf{Script Validation:} The dataset is rigorously validated to ensure linguistic and acoustic consistency before model training proceeds.
\end{enumerate}

\vspace{0.5em}
\textbf{Mel-Spectrogram Generation: }
Once training is completed, the Tacotron 2 model is employed to generate mel-spectrograms from new Meitei Mayek text inputs. These spectrograms represent the time-frequency characteristics of the expected audio output.

\vspace{0.5em}
\textbf{Speech Reconstruction Using HiFi-GAN Vocoder: }
The mel-spectrograms are subsequently passed through a HiFi-GAN vocoder, a generative adversarial network specialized in waveform reconstruction. The vocoder transforms the spectrograms into natural, intelligible speech signals with high fidelity.\\
Manual Dataset Review: After initial synthesis, manual reviews are conducted to detect and correct any perceptible errors, ensuring the final outputs meet quality standards.

\vspace{0.5em}
\textbf{Speech Output Generation}
The output of the vocoder stage constitutes the synthesized speech corresponding to the input Meitei Mayek text. The system thus achieves end-to-end text-to-speech conversion.

\vspace{0.5em}
\textbf{Evaluation and Iterative Refinement}
The performance of the TTS system is evaluated using both quantitative metrics (such as training and validation loss curves) and qualitative assessments (such as listening tests). Based on feedback, further refinements are made to enhance model performance and synthesis quality.

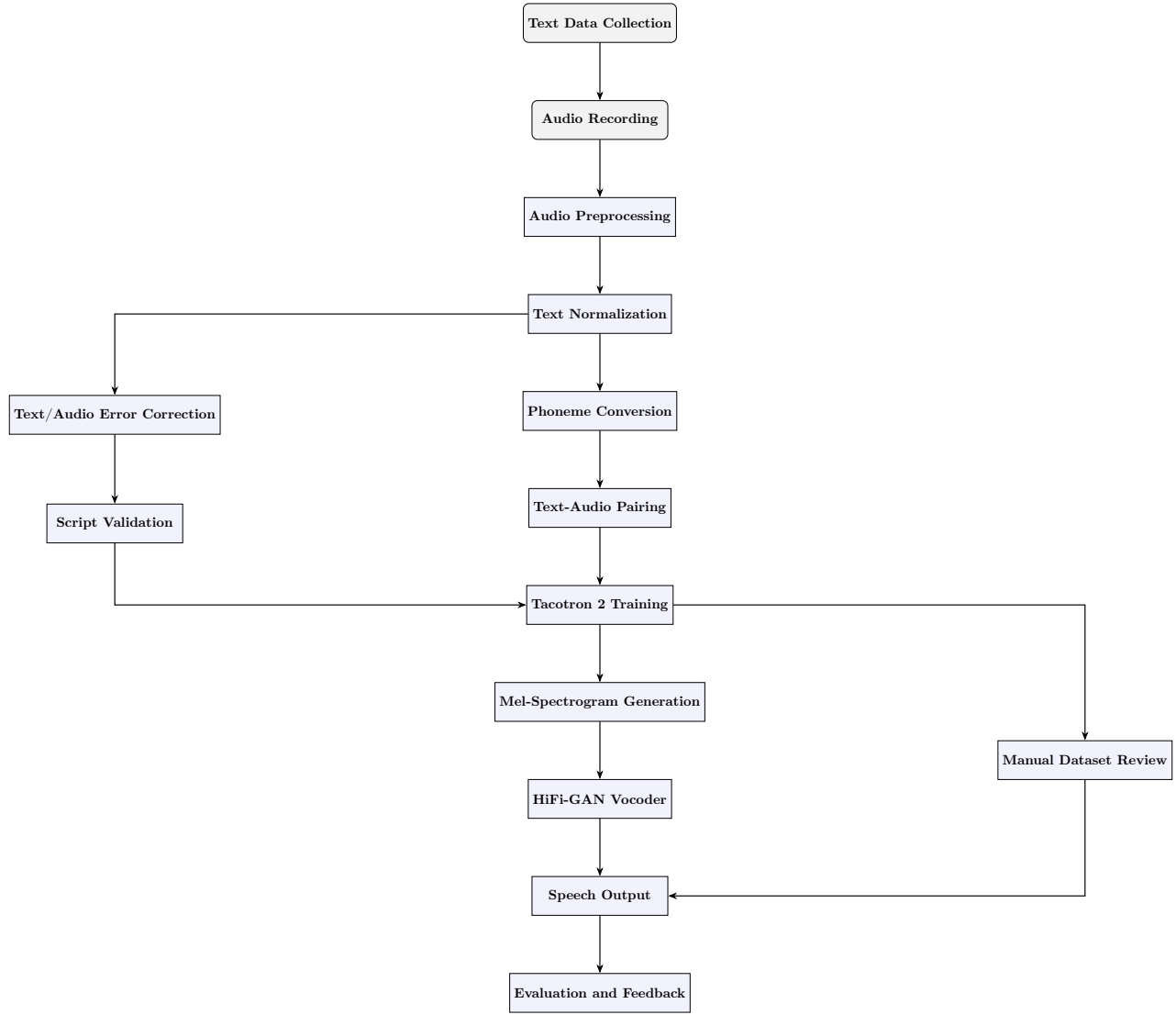
\begin{figure}[!ht]
    \begin{center}
        \begin{adjustbox}{max height=0.8\textheight, center, width=0.8\textwidth}
            \begin{tikzpicture}[node distance=2.5cm]
                \node (t0) [startstop] {\textbf{Text Data Collection}};
                \node (t1) [startstop, below of = t0] {\textbf{Audio Recording}};
                \node (t2) [process, below of = t1] {\textbf{Audio Preprocessing}};
                \node (t3) [process, below of = t2] {\textbf{Text Normalization}};
                \node (t4) [process, below of = t3] {\textbf{Phoneme Conversion}};
                \node (t5) [process, below of = t4] {\textbf{Text-Audio Pairing}};
                \node (t6) [process, below of = t5] {\textbf{Tacotron 2 Training}};
                \node (t7) [process, below of = t6] {\textbf{Mel-Spectrogram Generation}};
                \node (t8) [process, below of = t7] {\textbf{HiFi-GAN Vocoder}};
                \node (t9) [process, below of = t8] {\textbf{Speech Output}};
                \node (t10) [process, below of = t9] {\textbf{Evaluation and Feedback}};

                \node (l1) [process, left of = t3, xshift=-10cm, yshift=-2.6cm] {\textbf{Text/Audio Error Correction}};
                \node (l2) [process, below of = l1, yshift=-0.3cm] {\textbf{Script Validation}};
                \draw [arrow] (t3) -| (l1);
                \draw [arrow] (l1) -- (l2);
                \draw [arrow] (l2) |- (t6);

                \node (r1) [process, right of = t6, xshift=10cm, yshift=-4cm] {\textbf{Manual Dataset Review}};
                \draw [arrow] (t6) -| (r1);
                \draw [arrow] (r1) |- (t9);

                \draw [arrow] (t0) -- (t1);
                \draw [arrow] (t1) -- (t2);
                \draw [arrow] (t2) -- (t3);
                \draw [arrow] (t3) -- (t4);
                \draw [arrow] (t4) -- (t5);
                \draw [arrow] (t5) -- (t6);
                \draw [arrow] (t6) -- (t7);
                \draw [arrow] (t7) -- (t8);
                \draw [arrow] (t8) -- (t9);
                \draw [arrow] (t9) -- (t10);
            \end{tikzpicture}
        \end{adjustbox}
    \end{center}
    \caption{Study methodology for the Meitei Mayek TTS system.}
    \label{fig:tts_methodology}
\end{figure}

\subsection{Models and Algorithms}
\subsubsection{Design of the Tacotron 2 Framework}

Tacotron 2 is a deep learning-based architecture that facilitates direct conversion of textual input into mel-spectrograms, which can subsequently be transformed into human-like speech through vocoders such as HiFi-GAN. The architectural pipeline comprises the following primary components:

\begin{itemize}
    \item An embedding layer that encodes characters into dense vectors
    \item An encoder with integrated attention to capture contextual relevance
    \item A decoder supported by a post-processing network to generate smooth spectrograms
\end{itemize}

\noindent The following Python implementation outlines the key structural elements of the Tacotron 2 model:

\begin{lstlisting}[language=Python]
class Tacotron2(nn.Module):
    def __init__(self, hparams):
        super(Tacotron2, self).__init__()
        self.embedding = nn.Embedding(hparams.n_symbols, hparams.symbols_embedding_dim)
        std = sqrt(2.0 / (hparams.n_symbols + hparams.symbols_embedding_dim))
        val = sqrt(3.0) * std
        self.embedding.weight.data.uniform_(-val, val)

        self.encoder = Encoder(hparams)
        self.decoder = Decoder(hparams)
        self.postnet = Postnet(hparams)
\end{lstlisting}

\begin{figure}[h]
    \centering
    \includegraphics[width=0.9\textwidth]{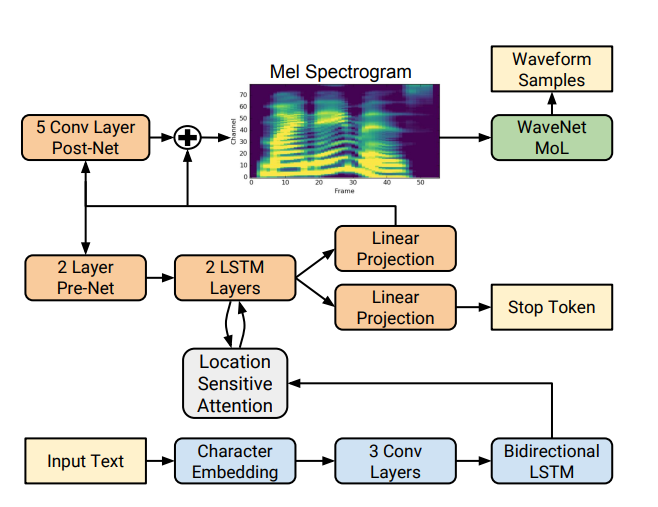}
\end{figure}

\subsubsection{Neural Network Models Used in Tacotron 2}

The Tacotron 2 architecture incorporates multiple types of neural networks, each suited for a particular role in the TTS pipeline:

\begin{itemize}

    \item \textbf{Prenet: } Serves as an information bottleneck and acts as a regularization module. It typically consists of \textbf{fully connected layers} with \textbf{dropout} and \textbf{ReLU} activation. It helps prevent overfitting and enhances generalization during both training and inference.
    
    \begin{itemize}
       \item The 2 Layer Pre-Net block immediately follows the LSTM layers in the decoder.
       \item It prepares input for the decoder by passing through fully connected layers with dropout and ReLU, acting as a regularizer.
    \end{itemize}

    \item \textbf{Encoder:} Utilizes 3 \textbf{Convolutional Neural Networks (CNNs)} to extract local sequential features from character embeddings. These are followed by a \textbf{Bidirectional Long Short-Term Memory (BiLSTM)} layer for capturing temporal dependencies in both directions.\\
    
    \quad Input Text $\rightarrow$ Character Embedding $\rightarrow$ 3 Convolutional Layers $\rightarrow$ Bidirectional LSTM

    \item \textbf{Decoder:} Composed of \textbf{stacked LSTM layers} that generate mel-spectrogram frames autoregressively, using prior outputs and attention context vectors.
    \textbf{Attention Mechanism:} Employs a \textbf{location-sensitive attention mechanism} powered by an \textbf{LSTM-based controller}. This mechanism aligns encoder outputs with decoder time steps using learned attention weights and prior positional information.\\
    \[
    \quad \text{Pre-Net} \rightarrow \text{2 LSTM Layers} \rightarrow \text{Linear Projection (Mel + Stop Token)}
   \]

     \begin{itemize}
       \item Location Sensitive Attention is looped into the decoder.
       \item It feeds attention-weighted encoder output into the LSTM decoder.
    \end{itemize}

    \item \textbf{Postnet:} Implements a stack of \textbf{1D CNN layers} to post-process and refine the predicted mel-spectrogram, resulting in smoother and higher-quality audio outputs.

      \begin{itemize}
       \item 5 Conv Layer Post-Net adds refinement via a residual connection.
       \item Output is then passed to vocoder for waveform synthesis.
    \end{itemize}
    
\end{itemize}

The Tacotron 2 model relies on several foundational neural network layers, each chosen for its ability to solve specific challenges in the speech synthesis pipeline. These layers operate in tandem to extract, transform, and model both spatial and temporal patterns in the input and output data.

\paragraph{Linear (Fully Connected) Layers :}

Linear layers, also referred to as dense layers, map an input vector to an output vector using a learnable weight matrix and bias vector. Given an input $\mathbf{x} \in \mathbb{R}^n$, the transformation is defined as:

\[
\mathbf{y} = W\mathbf{x} + \mathbf{b}
\]

Here, $W \in \mathbb{R}^{m \times n}$ is the weight matrix, $\mathbf{b} \in \mathbb{R}^m$ is the bias, and $\mathbf{y} \in \mathbb{R}^m$ is the result. In Tacotron 2, such layers appear in the Pre-net and decoder, where they help convert input features into forms that are easier for the subsequent modules to process.

\paragraph{ReLU (Rectified Linear Unit) :}

The ReLU activation function introduces non-linearity after linear transformations and helps mitigate the vanishing gradient issue in deep networks. Mathematically, it is defined as:

\[
\text{ReLU}(x) = \max(0, x)
\]

This function retains positive values while setting negative values to zero, allowing networks to learn sparse and efficient representations.

\paragraph{Dropout :}

Dropout is a form of regularization that randomly disables a portion of the neurons during training, thereby forcing the network to rely on multiple pathways. Given a dropout probability $p$, the output of each unit becomes:

\[
\tilde{x}_i =
\begin{cases}
0, & \text{with probability } p \\
\frac{x_i}{1 - p}, & \text{otherwise}
\end{cases}
\]

This technique, commonly used in the Pre-net, helps improve generalization by reducing overfitting.

\paragraph{Convolutional Layers :}

Convolutional layers apply sliding filters across the input sequence, enabling the network to identify local features that occur in spatially or temporally adjacent positions. A one-dimensional convolution operation can be expressed as:

\[
y_t = \sum_{i=1}^{k} w_i \cdot x_{t + i - \lfloor k/2 \rfloor}
\]

where $k$ is the filter width, $w_i$ are the learnable kernel weights, and $x$ is the input sequence. Tacotron 2 employs convolutional layers in the encoder and post-processing network to capture patterns in both text and spectrogram domains.

\paragraph{LSTM (Long Short-Term Memory) :}

LSTM units are specialized forms of recurrent networks designed to retain long-term dependencies. They use internal gating mechanisms to control the flow of information. The computations for each time step $t$ are:

\[
\begin{aligned}
f_t &= \sigma(W_f x_t + U_f h_{t-1} + b_f) \\
i_t &= \sigma(W_i x_t + U_i h_{t-1} + b_i) \\
o_t &= \sigma(W_o x_t + U_o h_{t-1} + b_o) \\
\tilde{c}_t &= \tanh(W_c x_t + U_c h_{t-1} + b_c) \\
c_t &= f_t \odot c_{t-1} + i_t \odot \tilde{c}_t \\
h_t &= o_t \odot \tanh(c_t)
\end{aligned}
\]

These equations define how the memory cell $c_t$ and the hidden state $h_t$ are updated. In Tacotron 2, LSTMs are used in the decoder and attention mechanism to model dependencies across time steps.

\paragraph{Bidirectional LSTM (BiLSTM) :}

BiLSTM networks enhance standard LSTMs by processing the input sequence in both forward and reverse directions. For a given time step $t$, the hidden representation is constructed by concatenating forward and backward hidden states:

\[
h_t = [\overrightarrow{h_t}; \overleftarrow{h_t}]
\]

This allows the model to utilize both past and future context when analyzing the input, which is particularly beneficial in the encoder where full context improves representation quality.

\paragraph{Attention Module with Recurrent Control :}

Tacotron 2’s attention system is implemented with an LSTM-based controller that guides the alignment between input tokens and output mel-spectrogram frames. At each decoding step, the attention module computes a context vector as a weighted sum over encoder outputs:

\[
\alpha_t = \text{Attention}(s_t, H)
\quad\text{and}\quad
c_t = \sum_{i} \alpha_{t,i} h_i
\]

where $s_t$ is the current decoder state, $H$ is the sequence of encoder outputs, $\alpha_t$ are the attention weights, and $c_t$ is the resulting context vector. This mechanism enables the model to dynamically focus on relevant parts of the input during synthesis.

\subsubsection{Embedding of Characters}

The use of embeddings in character-level TTS allows for efficient transformation of discrete tokens into meaningful numerical vectors. This is particularly important for languages with unique orthographies, such as Meitei Mayek.

\textbf{Theoretical Formulation}

Let the variables be defined as follows:

\begin{itemize}
    \item $V$ represents the total number of distinct characters (vocabulary size)
    \item $d$ denotes the embedding space dimension
    \item $\mathbf{x}_i$ is the one-hot representation of the $i^{th}$ character
    \item $W_{\text{emb}} \in \mathbb{R}^{V \times d}$ is the learnable embedding matrix
\end{itemize}

The embedded output vector is then calculated by:

\[
\mathbf{e}_i = W_{\text{emb}} \cdot \mathbf{x}_i
\]

where $\mathbf{e}_i$ is the resulting dense vector of size $d$.

\noindent\textbf{Implementation Snippet:}
\begin{lstlisting}[language=Python]
self.embedding = nn.Embedding(hparams.n_symbols, hparams.symbols_embedding_dim)
\end{lstlisting}

\subsubsection{Phonetic Mapping for Meitei Mayek}

To tailor the model for Meitei Mayek, each character is mapped to its approximate ARPAbet phonetic representation using a curated lookup table. This ensures phoneme-level granularity in the training data.

\subsubsection{Training Dynamics and Learning Rate Control}

During model training, a decaying exponential function is utilized to adaptively adjust the learning rate. This helps stabilize convergence and reduce overfitting.

\[
\text{LR} = \max\left(C, A \cdot e^{-\frac{\text{iteration} - \text{decay\_start}}{B}}\right)
\]

Where:
\begin{itemize}
    \item $A$ denotes the initial learning rate
    \item $B$ controls the rate of decay
    \item $C$ defines the minimum threshold
\end{itemize}

\subsubsection{Audio Synthesis Using HiFi-GAN}
After spectrograms are generated, a neural vocoder — HiFi-GAN — synthesizes realistic waveforms. This vocoder leverages adversarial training for enhanced audio fidelity and low inference latency.

\section{Data Collection and Extraction}
\subsection{Introduction}
\noindent

Data collection and extraction are the most important steps in building a Text-to-Speech (TTS) system. A TTS system can work properly only if it is properly trained with good-quality data mapping written text to speech. In this project, focus was on the Meitei language written in the Meitei Mayek script. The goal was to gather both written text and native speaker recordings to train the TTS system.

\subsubsection{Survey Site Selection}

For recording natural and accurate language data, we first selected appropriate survey locations. These locations were where Meitei Lon was used widely, such as valley regions in Manipur. Greater importance was placed on communities that used Meitei Lon as a common medium of everyday conversation. For this particular project we take assistance from a student in NIT, Manipur who is fluent in Meitei lon. This helped in ensuring the speech data that was recorded reflected the natural pronunciation and intonation of the language.

\subsubsection{Audio Data Collection and Preprocessing}

A native speaker is selected who is fluent in Manipuri and who is also available nearby. The speaker reads the Meitei Mayek script gathered from various sources like newspapers, text books etc. The audio Files are then preprocessed as follows:

\begin{itemize}
    \item \textbf{Silence Removal:} Unwanted silences are removed from within the audio clips. This is done so that the unwanted informations does not disrupt the training of the model.
    \item \textbf{Volume Normalization:} Audio clips are normalized to ensure the audio clips are equally audible. This ensure that the output speech  of the model is uniform.
    \item \textbf{Text and Audio Alignment:} The text is precisely aligned with its corresponding audio. This ensures that our model learns accurate mapping.
\end{itemize}

\subsubsection{Creating the Database}

To train the model for Manipuri language, we create a custom-built dataset consisting of native Manipuri speech and manually aligned
ARPAbet phoneme sequences. To accommodate this, we built a phoneme mapping scheme that relates Meitei Mayek characters with ARPAbet phonemes.

Finally, the text and audio pairs are organized into a database that will be used to train the
 Tacotron model. The data is structured in a way that makes it easy to match each audio
 clip with the corresponding text from the OCR process.

\section{Analysis and Results}
\subsection{Introduction}
Here, we examine more closely how the Meitei Mayek Text-to-Speech (TTS) system was designed and how effectively it worked. We test the system with a combination of quantitative figures and real-world listeners' feedback. As we do this, we'll point out what did and didn't work, and what we can learn from these results about the system's success in converting written Meitei text to natural-sounding speech. This breakdown provides a better understanding of the potential and limitations of our chosen approach, particularly for a language and a script that, at present, do not yet have many electronic tools.

\subsection{Descriptive Statistics}

For training and testing the TTS system, a text-aligned audio corpus in Meitei Mayek was used. The corpus was read to incorporate a variety of phonemes, sentence patterns, and linguistic expressions used in the Meitei language.

Some essential characteristics of the dataset are listed in Table~\ref{tab:dataset_stats}:

\begin{table}[h!]
\centering
\caption{Dataset Statistics for Meitei Mayek TTS System}
\label{tab:dataset_stats}
\begin{tabular}{|l|c|}
\hline
\textbf{Parameter} & \textbf{Value} \\
\hline
Total number of speech samples & 818 \\
Average characters per sentence & 6 \\
Number of unique Meitei Mayek characters & 55 \\
Total duration of audio recordings & 40 minutes \\
Audio sampling rate & 22.05 kHz \\
\hline
\end{tabular}
\end{table}

 For training and testing, the corpus was split into training, validation, and test sets in the ratio of 80:10:10. Averaged effort was used to get an even balance of phoneme and sentence lengths over the splits for better generalization and preventing overfitting while training.

\subsection{Model Development}

Our TTS system is made up of three-part design.

\subsubsection{Text Preprocessing}

The Meitei Mayek is first processed by cleaning and standardization to remove inconsistencies. It is then analyzed into basic sound units (phonemes), which helps the model understand how to pronounce each character.

\subsubsection{Acoustic Modeling}

We used a model called Tacotron 2 to translate the text (in phoneme form) to what is referred to as a Mel-spectrogram—a visual representation of how the speech is to be produced. This part of the model was then trained on a loss function comparing predicted outputs to recorded speech.

\subsubsection{Waveform Generation (Vocoder)}

Here we use a vocoder called Hifi-GAN to convert spectrogram into actual speech audio. Hifi-Gan is more widely used for it's efficiency and ability to produce clearer natural speech.

\subsubsection{Training Details}

The training setup for the model is as follows:

\begin{itemize}
    \item \textbf{Optimizer:} Adam 
    \item \textbf{Initial Learning Rate:} 0.001 with exponential decay schedule.
    \item \textbf{Total Epochs:} 310
    \item \textbf{Batch Size:} 32
    \item \textbf{Convergence:} The model shows a steady decline in training loss across epochs, with significant reduction during the first 150 epochs. After a brief spike near epoch 180, the loss stabilizes and gradually converges, maintaining a value below 0.08 by the final epochs, indicating effective learning and convergence.
\end{itemize}

\subsubsection{Evaluation Metrics}

The model's performance was assessed using both objective and subjective evaluation metrics:

\begin{itemize}
    
    \item \textbf{Mean Opinion Score (MOS):} 
    \begin{itemize}
    \item \textbf{Naturalness:} 3.34 +- 0.70
    \item \textbf{Pronunciation Accuracy:} 3.51 +- 0.31
    \item \textbf{Overall:} 3.43 +- 0.51
    \end{itemize}

\end{itemize}

These results suggest that the system is capable of generating intelligible and fairly natural-sounding speech, despite the challenges associated with low-resource language processing.

\subsection{Discussion}
As we can see from the results, the performance of our Text-To-Speech is quite encouraging, despite the lack of availability of resources for the language. The generated speech
outputs were understandable and clear, and in a majority of the cases, natural into-
nation and rhythm were exhibited by the system, especially with short and popular
words.

However, several limitations were observed during evaluation:
\begin{itemize}
    \item  Some combination of phoneme sequence which is rarely present in the training dataset were not pronounced accurately. This may be due to lack of exposure when training the model.
    
    \item  When encountering long and syntactically complex sentences the generated audio sometimes show lack of natural rhythm or sounded too flat.
    
    \item  Certain tonal and stress characteristics unique to the Meitei language were not well preserved by the model, possibly due to the lack of explicit prosody modeling or annotated prosody information in the dataset.
\end{itemize}

Despite these limitations, according to our surveys, which consists of mostly native speakers, gave a relatively positive review saying that the speech was understandable in most cases. These results suggest that our model has learned the fundamental phonetics and rhythm patterns of the language relatively well.

Here, improvements can be done such as, increasing the dataset size and applying linguistic preprocessing could further increase the quality of output of this system. Additionally, by exploring transfer learning from models which is trained on other related models may be beneficial for improving the pronunciation and rhythm in future works.

\section{Conclusions and Practical
Implications}
The objective of this research was to develop the world's first-ever Text-to-Speech system for Manipuri language written in the Meitei Mayek script. This project is technologically as well as culturally significant. Most contemporary TTS systems are based on a small number of popular languages and scripts, but Meitei Mayek is hardly visible in the digital sphere. This project attempts to fill that gap. This research demonstrates
emonstrated that natural-sounding speech can be generated from Meitei Mayek
script using the design of a phoneme set emulating the Manipuri language and
adapting neural TTS models like Tacotron 2 and HiFi-GAN. The system yielded
promising results considering this is a first attempt and is sure to be helpful
in many fields like accessibility, education, and digital inclusion.

\subsection{Study Findings}
Some of the most important findings from this research were:

First, a personalized mapping of Meitei Mayek characters to their phonetic sounds
i.e ARPAbet phonemes was created. This enabled the model to process the native
script directly without having to transliterate it into English or another language
first.

The Tacotron 2 model, having been trained on Manipuri audio data and augmented with
HiFi-GAN, generated speech that was clean and intelligible. Even with training data being fairly small,
the system was able to generate output that very much resembled natural human speech.

The results confirmed that an optimally designed set of phonemes and an enlightened adaptation of existing deep learning models can overcome a lot of the limitations that low-
resource languages typically find in speech technology.

\subsection{Research Contributions}
This work has the following new contributions:

Phoneme Design: The largest contribution is designing a Meitei Mayek phoneme set for use in TTS. There was no such tool available previously and it is a foundational block for any future language tech tools in Manipuri.

TTS Adaptation: Training a strong TTS model such as Tacotron 2 on Meitei Mayek-written text, but aligning its phonemes with the ARPAbet notation, illustrates a new approach to crossing indigenous scripts with popular speech synthesis platforms.

New Dataset: The team developed and utilized a parallel dataset of Meitei Mayek text and Manipuri speech. The corpus could highly be useful to other developers and researchers in similar fields.

Proof of Concept for Underresourced Languages: Most significantly, the project demonstrates that highly underresourced languages like Manipuri can be empowered with existing AI tools if one thinks outside the box.

\subsection{Practical Implications}

Practical uses for this work are of wide ranges and has a deeply impact:

Accessibility: This TTS system has the potential to be of tremendous help for visually impaired individuals or persons with reading disabilities. They can listen rather than read because they will have access to content presented in Meitei Mayek.

Learning and Education: Students learning Manipuri can listen to correct pronunciations, which enable them to enhance their spoken skill. Teachers and developers can leverage this system within language learning apps to enhance learning.

Cultural Conservation: Manipuri script and language have a rich history behind them. We can translate folk tales, stories and other forms of cultural works by trans-
forming them into audio.

Public Communication: he government and public administrations are in a position to install this technology to disseminate information in the Manipuri language so that people can be able to obtain necessary
supply updates in their own language and form..

Media and Content Creation: The platform can further be used to produce audio
books, podcasts, and local content, encouraging more digital content in Meitei Mayek.We can convert folk tales, stories and other types of cultural works by con-
verting them into audio.

\subsection{Study Limitations and Scope for Future Work}
While this project was able to accomplish its principal objectives, there are a few
limits which must be highlighted:

Limited Data and Speaker Variety: Training was done with only a single speaker’s
data.To that end, it does not have voice diversity, accent diversity, and variety of
styles.A variety of speakers would make more flexible and spontaneous output.

Lack of Expressive Speech: The present system is not ready to cope with variability in speaking like empha-
sis, tone, or emotion. They are of most significance in synthesizing speech as
as natural as can be. Future work may be directed towards incorporating those features-
tures.. Simple Evaluation Tools: Though we tried the system with simple tools such
as MOS scores and spectrogram comparison, even more analysis on native speaker
feedback would yield better understanding.

Evaluative Methods: The system’s architecture is optimized for laboratory use only
and can be used on embedded systems with moderate improvement to provide
real-time operation; nonetheless, it will not work on mobile phones unless with extensive
changes.

Assessment Procedures: The system testing approach did not involve comprehensive
qualitative inspection of the audio data gathered, and it depended on simple methods like MOS scoring and spectrogram analysis. Analysis based on interactions with
native speakers would significantly refine understanding.

Not Yet Optimized for Real-Time Use: The system currently functions in a lab setup
but would need further development to comfortably function on embedded systems
or smartphones for real-time usage.

Potential for Broader Applications: The approach adopted here can be taken to
other less-researched Indian scripts or tribal languages. Expanding this research to
new environments, we can create more diverse technology for other individuals as
well.

\section*{Acknowledgements}

We would like to express our deepest gratitude to our supervisor, Dr. Khundrakpam Johnson Singh, and to Dr. Chingakham Neeta Devi for their invaluable guidance, support, and encouragement throughout the course of our project.

We are also sincerely thankful to Ms. Thokchom Feemi Devi for her valuable contribution in providing the voice recordings that helped complete our dataset.

\nocite{*}
\label{References}

\lhead{\emph{References}} 
\renewcommand\bibname{References}
\bibliographystyle{unsrt}
\bibliography{references} 

\end{document}